\documentclass[11pt,a4paper]{article}
\usepackage[hyperref]{acl2020}
\usepackage{times}
\usepackage{latexsym}
\usepackage{helvet}
\usepackage{subfigure}
\usepackage{courier}
\usepackage{bm}
\usepackage{amsfonts}
\usepackage{amsmath}
\usepackage{url}
\usepackage{graphicx}
\usepackage{siunitx}
\usepackage{booktabs}
\usepackage{multirow}
\usepackage{enumitem}
\setitemize[1]{leftmargin=*,parsep=0pt}

\usepackage{microtype}

\aclfinalcopy 

\title{{T}ext{B}rewer: {A}n {O}pen-{S}ource {K}nowledge {D}istillation {T}oolkit for {N}atural {L}anguage {P}rocessing}

\author{
Ziqing Yang$^\dag$,
Yiming Cui$^\ddag$$^\dag$,
Zhipeng Chen$^\dag$, \\
\textbf {
Wanxiang Che$^\ddag$,
Ting Liu$^\ddag$,
Shijin Wang$^\dag$$^\S$,
Guoping Hu$^\dag$
}\\
{$^\dag$State Key Laboratory of Cognitive Intelligence, iFLYTEK Research, China} \\
{$^\ddag$Research Center for Social Computing and Information Retrieval (SCIR), } \\
{Harbin Institute of Technology, Harbin, China} \\
{$^\S$iFLYTEK AI Research (Hebei), Langfang, China} \\
$^\dag$$^\S$\tt\{zqyang5,ymcui,zpchen,sjwang3,gphu\}@iflytek.com \\
$^\ddag$\tt\{ymcui,car,tliu\}@ir.hit.edu.cn}

\date{}

\begin{document}
\maketitle
\begin{abstract}
In this paper, we introduce {\bf TextBrewer}, an open-source knowledge distillation toolkit designed for natural language processing.
It works with different neural network models and supports various kinds of supervised learning tasks, such as text classification,  reading comprehension, sequence labeling.
TextBrewer provides a simple and uniform workflow that enables quick setting up of distillation experiments with highly flexible configurations.
It offers a set of predefined distillation methods and can be extended with custom code.
As a case study, we use TextBrewer to distill BERT on several typical NLP tasks.
With simple configurations, we achieve results that are comparable with or even higher than the public distilled BERT models with similar numbers of parameters. \footnote{TextBrewer: \url{http://textbrewer.hfl-rc.com}}

\end{abstract}

\section{Introduction}
Large pre-trained language models, such as GPT \cite{gpt}, BERT \cite{devlin-etal-2019-bert}, RoBERTa \cite{DBLP:journals/corr/abs-1907-11692} and  XLNet \cite{DBLP:journals/corr/abs-1906-08237}
 have achieved great success in many NLP tasks and greatly contributed to the progress of NLP research. 
However, one big issue of these models is the high demand for computing resources ---
they usually have hundreds of millions of parameters, and take several gigabytes of memory to train and inference --- 
which makes it impractical to deploy them on mobile devices or online systems. 
From a research point of view, we are tempted to ask: 
is it necessary to have such a big model that contains hundreds of millions of parameters to achieve a high performance? 
Motivated by the above considerations, recently, some researchers in the NLP community have tried to design lite models \cite{albert}, or resort to knowledge distillation (KD) technique
to compress large pre-trained models to small models.

KD is a technique of transferring knowledge from a teacher model to a student model, which is usually smaller than the teacher.
The student model is trained to mimic the outputs of the teacher model. 
Before the birth of BERT, KD had been applied to several specific tasks like machine translation \cite{kim-rush-2016-sequence, DBLP:conf/iclr/TanRHQZL19} in NLP. 
While the recent studies of distilling large pre-trained models focus on finding general distillation methods that work on various tasks and are receiving more and more attention
 \cite{sanh2019distilbert, DBLP:journals/corr/abs-1909-10351, sun-etal-2019-patient, tang-etal-2019-natural, DBLP:journals/corr/abs-1904-09482, clark-etal-2019-bam, DBLP:journals/corr/abs-1909-11687}.

Though various distillation methods have been proposed, they usually share a common workflow: 
firstly, train a teacher model, then optimize the student model by minimizing some losses that are calculated between the outputs of the teacher and the student.
Therefore it is desirable to have a reusable distillation workflow framework and treat different distillation strategies and tricks as plugins so that they could be easily and arbitrarily added to the framework.
In this way, we could also achieve great flexibility in experimenting with different combinations of distillation strategies and comparing their effects.

In this paper, we introduce {\bf TextBrewer}, a PyTorch-based distillation toolkit for NLP
that aims to provide a unified distillation workflow, save the effort of setting up experiments and help users to distill more effective models.
TextBrewer provides simple-to-use APIs, a collection of distillation methods, and highly customizable configurations. 
It has also been proved able to distill BERT models efficiently and reproduce the state-of-the-art results on typical NLP tasks.
The main features of TextBrewer are:

\begin{itemize}
\item \textbf{Versatility in tasks and models}. It works with a wide range of models, from the RNN-based model to the transformer-based model,
and works on typical natural language understanding tasks.
Its usability in tasks like text classification, reading comprehension, and sequence labeling has been fully tested.

\item \textbf{Flexibility in configurations}. 
The distillation process is configured by configuration objects, which can be initialized from JSON files and contain many tunable hyperparameters.
Users can extend the configurations with new custom losses, schedulers, etc., if the presets do not meet their requirements.

\item \textbf{Including various distillation methods and strategies}. 
KD has been studied extensively in computer vision (CV) and has achieved great success.
It would be worthwhile to introduce these studies to the NLP community as some of the methods in these studies could also be applied to texts. 
TextBrewer includes a set of methods from both CV and NLP,
such as  flow of solution procedure (FSP) matrix loss \cite{DBLP:conf/cvpr/YimJBK17}, neuron selectivity transfer (NST) \cite{DBLP:journals/corr/HuangW17a}, probability shift and dynamic temperature \cite{DBLP:journals/corr/abs-1911-07471}, attention matrix loss, multi-task distillation \cite{DBLP:journals/corr/abs-1904-09482}. 
In our experiments, we will show the effectiveness of applying methods from CV on NLP tasks.

\item \textbf{Being non-intrusive and simple to use}. 
\emph{Non-intrusive} means there is no need to modify the existing code that defines the models.
Users can re-use the most parts of their existing training scripts, such as model definition and initialization, data preprocessing and task evaluation.  Only some preparatory work (see Section \ref{Workflow}) are additionally required to use TextBrewer to perform the distillation.

\end{itemize}

TextBrewer also provides some  useful utilities such as model size analysis and data augmentation
to help model design and distillation.

\section{Related Work}
Recently some distilled BERT models have been released, such as DistilBERT \cite{sanh2019distilbert}, TinyBERT \cite{DBLP:journals/corr/abs-1909-10351}, and ERNIE Slim\footnote{{https://github.com/PaddlePaddle/ERNIE}}.
DistilBERT performs distillation on the pre-training task, i.e., masked language modeling. 
TinyBERT performs transformer distillation at both the pre-training and task-specific learning stages. 
ERNIE Slim distills ERNIE \cite{baidu-ernie-1,baidu-ernie-2}on a sentiment classification task.
Their distillation code is publicly available, and users can replicate their experiments easily.
However, it is laborious and error-prone to change the distillation method or adapt the distillation code for some other models and tasks,  
since the code is not written for general distillation purposes. 

There also exist some libraries for general model compression.
Distiller \cite{neta_zmora_2018_1297430} and PaddleSlim\footnote{{https://github.com/PaddlePaddle/PaddleSlim}} are two versatile libraries supporting pruning, quantization and knowledge distillation. 
They focus on models and tasks in computer vision.
In comparison, TextBrewer is more focused on knowledge distillation on NLP tasks, more flexible, and offers more functionalities. 
Based on PyTorch, It provides simple APIs and rich customization for fast and clean implementations of experiments.

\begin{figure*}[!tbp]
 \centering
 \subfigure[]{
  \includegraphics[width=0.63\textwidth]{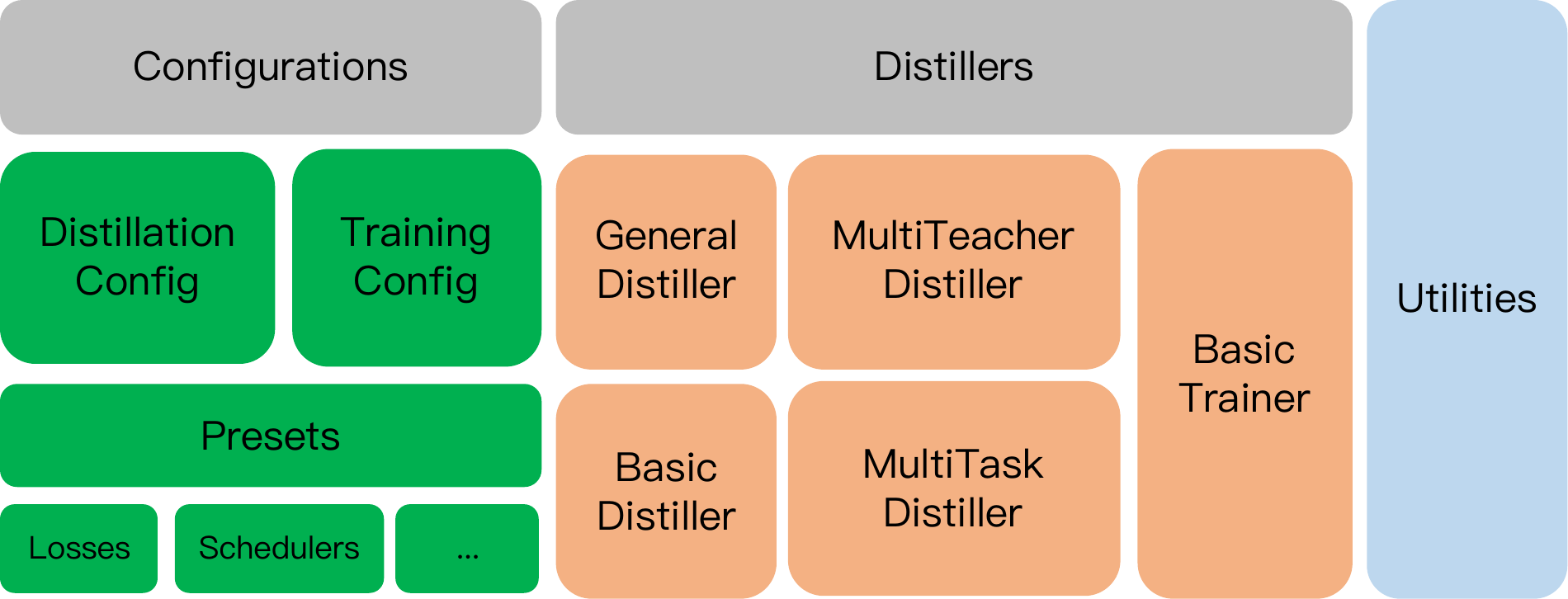} 
 }
 \subfigure[]{
  \includegraphics[width=0.30\textwidth]{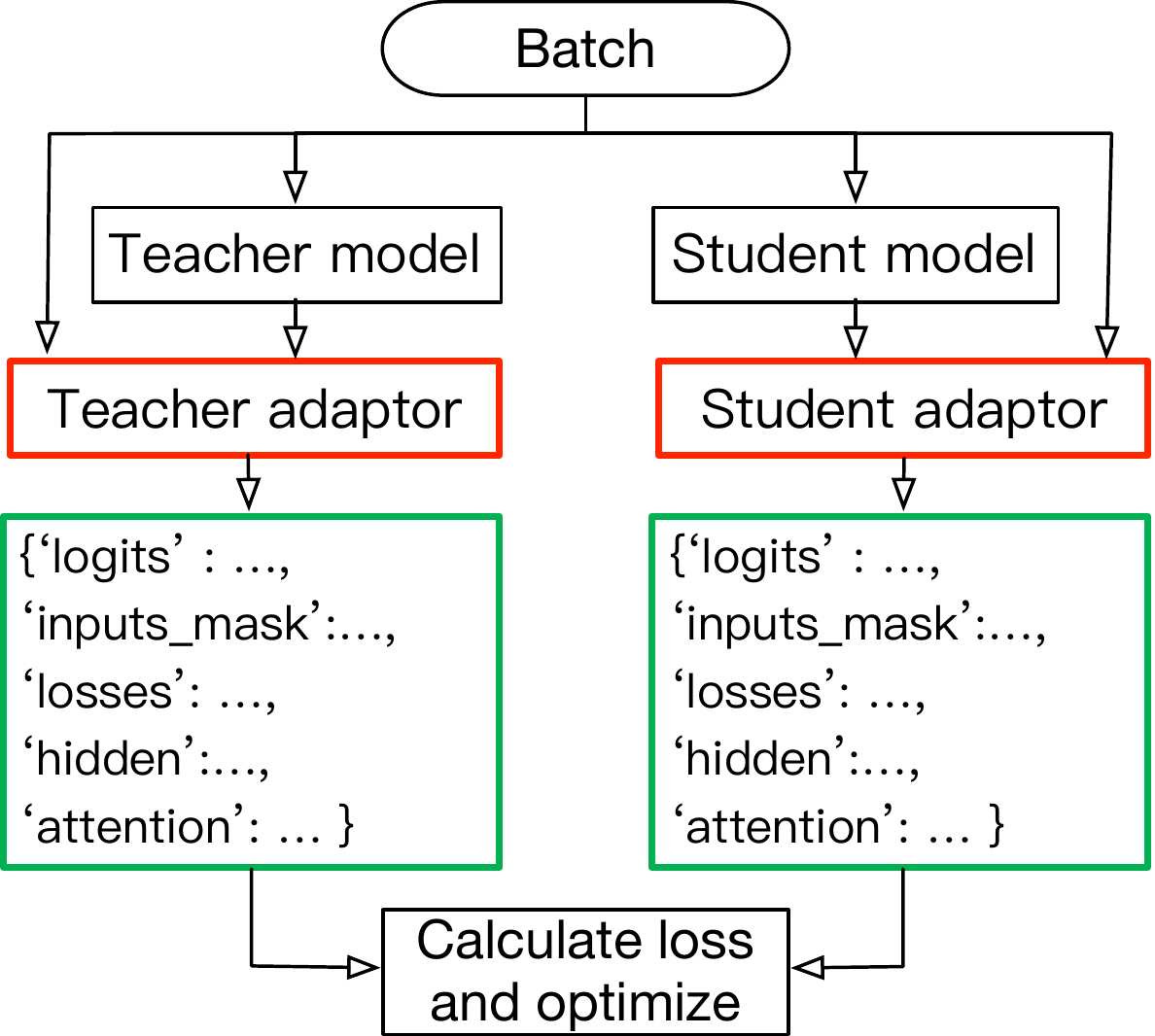} 
 }
 \caption{\label{overview} (a) An overview of the main functionalities of TextBrewer. (b) A sketch that shows the function of adaptors inside a distiller.}
\end{figure*}

\section{Architecture and Design} 

Figure \ref{overview} shows an overview of the main functionalities and architecture of TextBrewer.
To support different models and different tasks and meanwhile stay flexible and extensible, 
TextBrewer provides \textit{distillers} to conduct the actual experiments and configuration classes to configure the behaviors of the distillers.

\subsection{Distillers}
Distillers are the cores of TextBrewer. They automatically train and save models and support custom evaluation functions.
Five distillers have been implemented: 
\texttt{BasicDistiller} is used for single-task single-teacher distillation; 
\texttt{GeneralDistiller} in addition supports more advanced intermediate loss functions;
\texttt{MultiTeacherDistiller} distills an ensemble of teacher models into a single student model;
\texttt{MultiTaskDistiller} distills multiple teacher models of different tasks into a single multi-task student model \cite{clark-etal-2019-bam, DBLP:journals/corr/abs-1904-09482}. 
We also have implemented \texttt{BasicTrainer} for training teachers on labeled data to unify the workflows of supervised learning and distillation.
All the distillers share the same interface and usage. They can be replaced by each other easily.

\subsection{Configurations and Presets}
The general training settings and the distillation method settings of a distiller are specified by two configurations: \texttt{TrainingConfig} and \texttt{DistillationConfig}.

\textbf{TrainingConfig} defines the settings that are general to deep learning experiments, including the directory where logs and student model are stored (\texttt{log\_dir}, \texttt{output\_dir}), the device to use (\texttt{device}), the frequency of storing and evaluating student model (\texttt{ckpt\_frequencey}), etc.
    
 \textbf{DistillationConfig} defines the settings that are pertinent to distillation, where various distillation methods could be configured or enabled. It includes the type of KD loss (\texttt{kd\_loss\_type}), the temperature and weight of  KD loss (\texttt{temperature} and \texttt{kd\_loss\_weight}), the weight of hard-label loss (\texttt{hard\_label\_weight}),  probability shift switch, schedulers and intermediate losses, etc. 
Intermediate losses are used for computing the losses between the intermediate states of teacher and student, 
and they could be freely combined and added to the distillers. Schedulers are used to adjust loss weight or temperature dynamically.

The available values of configuration options such as loss functions and schedulers are defined as dictionaries in presets.
For example, the loss function dictionary includes hidden state loss, cosine similarity loss, FSP loss, NST loss, etc.

\begin{figure}[htbp]
\centering
\includegraphics[width=\columnwidth]{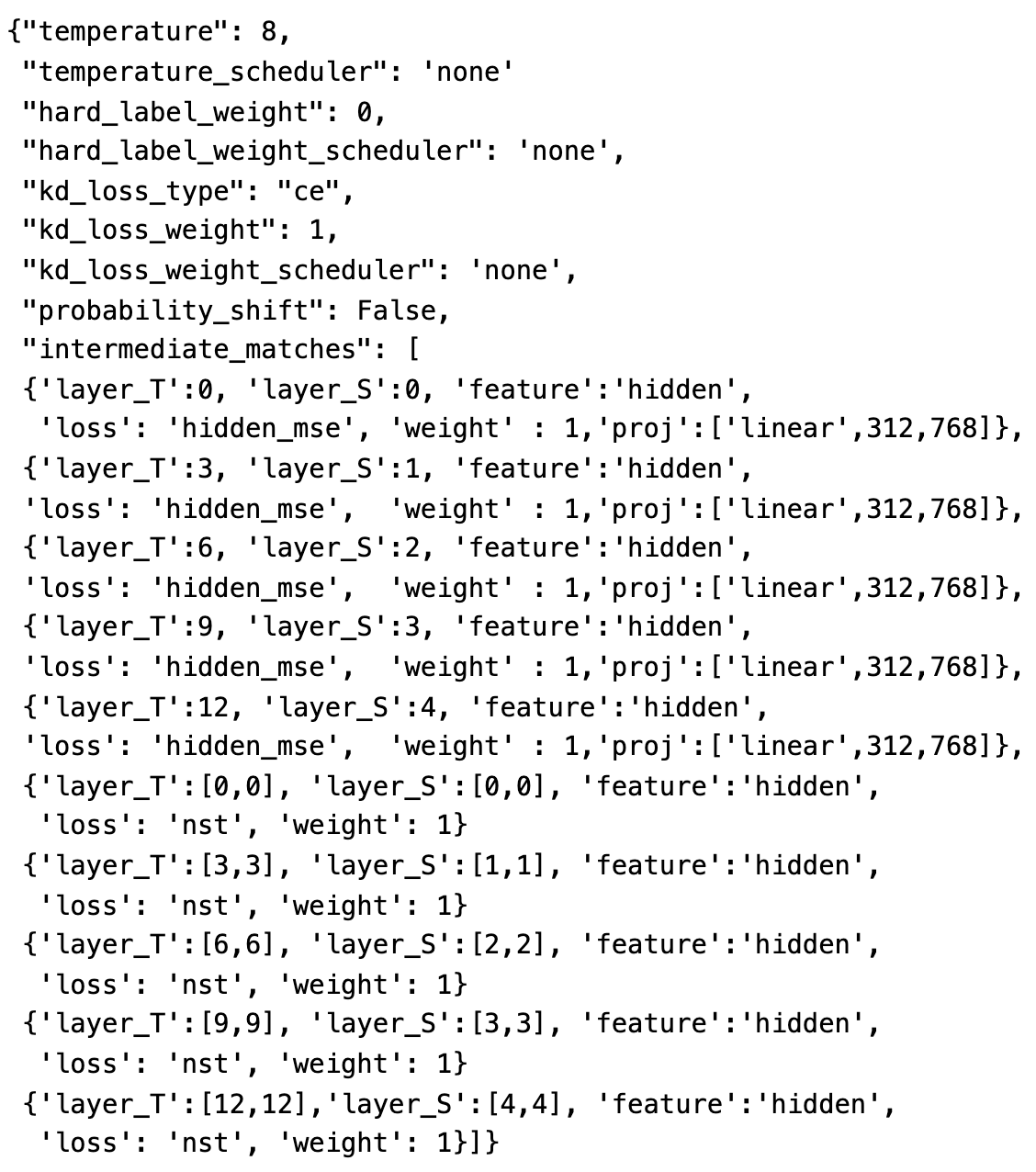}
\caption{\label{distill-config} An example of distillation configuration. This configuration is used to distill a 12-layer BERT$_{\text{\tt BASE}}$ to a 4-layer T4-tiny.}
\end{figure}

All the configurations can be initialized from JSON files.
In Figure \ref{distill-config} we show an example of \texttt{DistillationConfig} for distilling BERT$_{\text{\tt BASE}}$, to a 4-layer transformers. See Section \ref{experiments} for more details.

\subsection{Workflow}\label{Workflow}

\begin{figure}[t]
\centering
\includegraphics[width=\columnwidth]{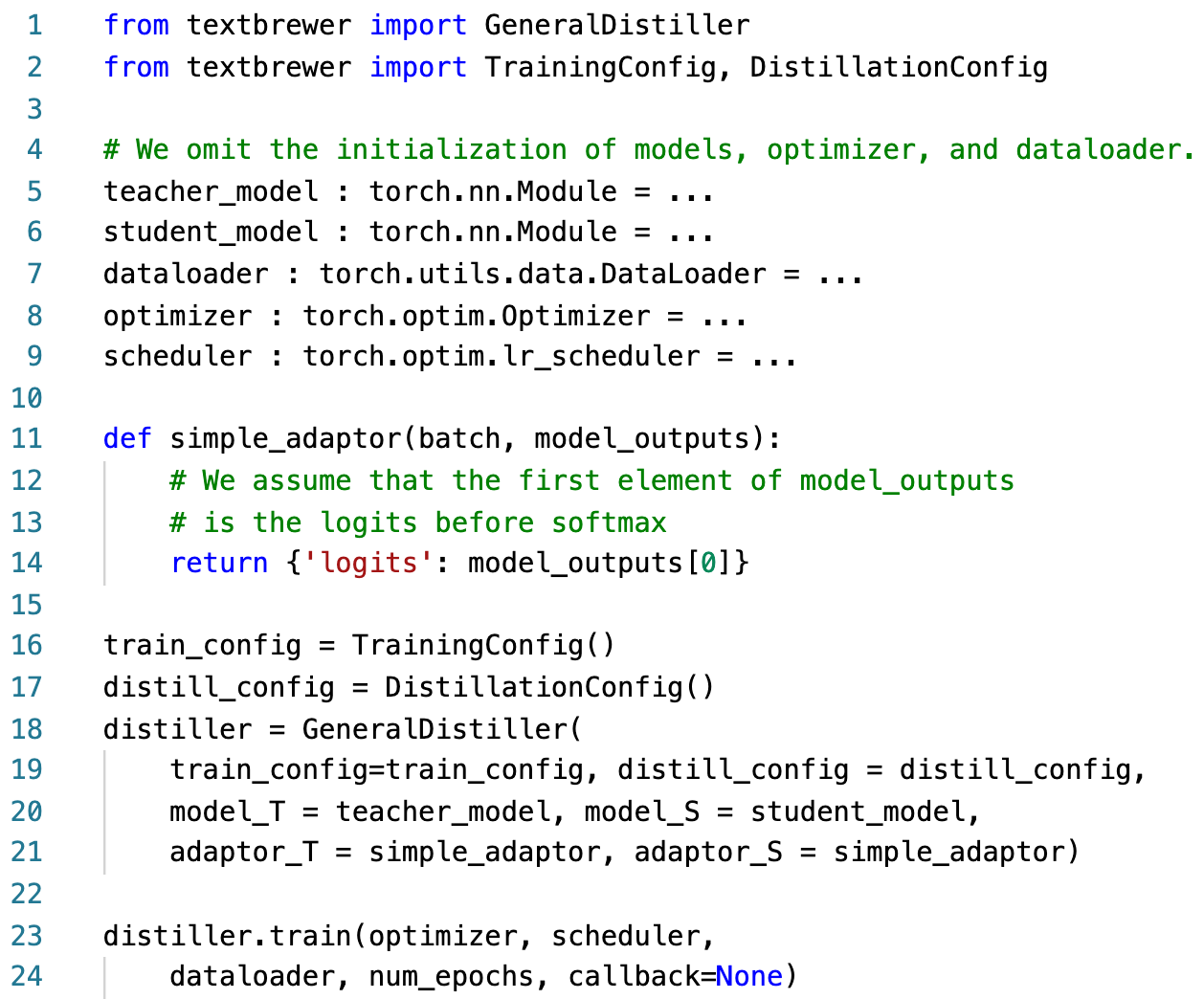}
\caption{\label{minimal} A  code snippet that demonstrates the minimal TextBrewer workflow.
}
\end{figure}

Before distilling a teacher model using TextBrewer, some preparatory works have to be done: 
\begin{enumerate}
\item Train a teacher model on a labeled dataset. Users usually train the teacher model with their own training scripts. 
    TextBrewer also provides \texttt{BasicTrainer} for supervised training on a labeled dataset.
\item Define and initialize the student model.
\item Build a dataloader of the dataset for distillation and initialize the optimizer and learning rate scheduler.
\end{enumerate}
The above steps are usually common to all deep learning experiments. 
To perform distillation, take the following additional steps:
\begin{enumerate}
    \item Initialize training and distillation configurations, and construct a distiller.
    \item Define \textit{adaptors} and a \textit{callback} function.
    \item Call the \texttt{train} method of the distiller.
\end{enumerate}
A code snippet that shows the minimal workflow is presented in Figure \ref{minimal}.
The concepts of callback and adaptor will be explained below.

\subsubsection{Callback Function}
To monitor the performance of the student model during training,  people usually evaluate the student model on a development set at some checkpoints besides logging the loss curve. 
For example, in the early stopping strategy, users choose the best model weights checkpoint based on the performance of the student model on the development set at the end of each epoch.
TextBrewer supports such functionality by providing the callback function argument in the \texttt{train} method, as shown in line 24 of Figure \ref{minimal}. The callback function takes two arguments: the student model and the current training step. 
At each checkpoint step (determined by \texttt{num\_train\_epochs} and \texttt{ckpt\_frequencey}), the distiller saves the student model and then calls the callback function.

Since it is impractical to implement evaluation metrics and evaluation procedures for all NLP tasks, we encourage users to implement their own evaluation functions as the callbacks for the best practice.

\subsubsection{Adaptor}
The distiller is model-agnostic. It needs a translator to translate the model outputs into meaningful data.
Adaptor plays the role of translator.
An Adaptor is an interface and responsible for explaining the inputs and outputs of the teacher and student for the distiller.

Adaptor takes two arguments: the model inputs and the model outputs. It is expected to return a dictionary with some specific keys.
Each key explains the meaning of the corresponding value, as shown in Figure \ref{overview} (b). 
For example, \texttt{logits} is the logits of final outputs, \texttt{hidden} is intermediate hidden states, \texttt{attention} is the attention matrices, \texttt{inputs\_mask} is used to mask padding positions. 
The distiller only takes necessary elements from the outputs of adaptors according to its distillation configurations. A minimal adaptor only needs to explain logits, as shown in lines 11--14 of Figure \ref{minimal}.

\subsection{Extensibility}
TextBrewer also works with users' custom modules. New loss functions and schedulers can be easily added to the toolkit. 
For example, to use a custom loss function, one first implements the loss function with a compatible interface,
then adds it to the loss function dictionary in the presets with a custom name, so that the new loss function becomes available as a new option value of the configuration and can be recognized by distillers.

\section{Experiments}\label{experiments}

In this section, we conduct several experiments to show TextBrewer's ability to distill large pre-trained models on different NLP tasks and achieve results are comparable with or even higher than the public distilled BERT models with similar numbers of parameters. \footnote{ More results are presented in the online documentation: \url{https://textbrewer.readthedocs.io}}

\subsection{Settings}

\textbf{Datasets and tasks.}
We conduct experiments on both English and Chinese datasets.
For English datasets, 
We use MNLI \cite{DBLP:conf/iclr/WangSMHLB19} for text classification task, SQuAD1.1 \cite{rajpurkar-etal-2016-squad} for  span-extraction machine reading comprehension (MRC) task and CoNLL-2003 \cite{tjong-kim-sang-de-meulder-2003-introduction} for named entity recognition (NER) task. 
For Chinese datasets, we use the Chinese part of XNLI \cite{conneau-etal-2018-xnli}, LCQMC \cite{liu-etal-2018-lcqmc}, CMRC 2018 \cite{cui-emnlp2019-cmrc2018} and DRCD \cite{DBLP:journals/corr/abs-1806-00920}. XNLI is the multilingual version of MNLI. LCQMC is a large-scale Chinese question matching corpus. CMRC 2018 and DRCD are two span-extraction machine reading comprehension datasets similar to SQuAD.
The statistics of the datasets are listed in Table \ref{datasets-statistics}.

\begin{table}[tbp]
\resizebox{\columnwidth}{!}{
\begin{tabular}{lcccc }
\toprule
\textbf{Dataset}& \textbf{Task} & \textbf{Metrics} & \textbf{\#Train} & \textbf{\#Dev} \\ \midrule
MNLI & Classification & Acc & 393K & 20K\\ 
SQuAD & MRC & EM/F1 & 88K &  11K \\
CoNLL-2003 & NER & F1 & 23K & 6K \\ \midrule
XNLI  & Classification & Acc & 393K & 2.5K \\
LCQMC & Classification & Acc & 293K & 8.8K \\
CMRC 2018 &  MRC & EM/F1 & 10K & 3.4K \\ 
DRCD &  MRC & EM/F1 & 27K & 3.5K \\ 
\bottomrule
\end{tabular}
}
\caption{\label{datasets-statistics}  A summary of the datasets used in experiments. The size of CoNLL-2003 is measured in number of entities.}
\end{table}

\textbf{Models.} All the teachers are BERT$_{\text{\tt BASE}}$-based models. 
For English tasks,  teachers are initialized with the weights released by Google\footnote{https://github.com/google-research/bert} and converted into PyTorch format via Transformers\footnote{https://github.com/huggingface/transformers}. 
For Chinese tasks, teacher is initialized with the pre-trained RoBERTa-wwm-ext \footnote{{https://github.com/ymcui/Chinese-BERT-wwm}} \cite{DBLP:journals/corr/abs-1906-08101}.
 We test the performance of the following student models:
  \begin{itemize}
 \item T6 and T3 are BERT$_{\text{\tt BASE}}$ with fewer layers of transformers. Especially, T6 has the same structure as DistilBERT \cite{sanh2019distilbert}.
 \item T3-small is a 3-layer BERT with half BERT-base's hidden size and feed-forward size.
 \item T4-tiny is the same as TinyBERT,  a 4-layer model with an even smaller hidden size and feed-forward size. 
 \item BiGRU is a single-layer bidirectional GRU. Its word embeddings are taken from BERT$_{\text{\tt BASE}}$. 
 \end{itemize}
T3-small and T4-tiny are initialized randomly. The model structures of the teacher and students are summarized in Table \ref{model-configurations}. 

\begin{table}[tbp]
\resizebox{\columnwidth}{!}{
    \begin{tabular}{lccccc}
    \toprule
    \multirow{2}{*}{\textbf{Model}} & \multicolumn{2}{c}{\textbf{MNLI}} & \multicolumn{2}{c}{\textbf{SQuAD}} & \textbf{CoNLL-2003}\\ 
     & m & mm & EM & F1 & F1        \\ \midrule
    BERT$_{\text{\tt BASE}}$  & 83.7 & 84.0 & 81.5 & 88.6  & 91.1 \\ \midrule
    \textit{Public} & & & & & \\
    DistilBERT & 81.6 & 81.1 & 79.1 & 86.9 & -\\
    TinyBERT & 80.5 & 81.0 &- & -  &- \\ 
    \ \ +DA & 82.8 & 82.9  &  72.7& 82.1 & - \\ \midrule
    \textit{TextBrewer} & & & & & \\
            BiGRU & - & - & - & - & 85.3 \\
    T6 & 83.6 & 84.0 & 80.8 & 88.1 & 90.7\\

    T3 & 81.6 & 82.5 & 76.3 & 84.8 & 87.5  \\
    T3-small & 81.3 & 81.7 & 72.3 & 81.4 & 78.6 \\
    T4-tiny  & 82.0 & 82.6 & 73.7 & 82.5 & 77.5 \\
    \ \ +DA & - & - & 75.2 & 84.0 & 89.1\\ 
\bottomrule
    
    \end{tabular}
    }
    \caption{\label{distiilation-results-english} Performance of BERT$_{\text{\tt BASE}}$ (teacher) and various students on the  development sets of MNLI and SQuAD, and the test set of CoNLL-2003. 
    \emph{m} and \emph{mm} under MNLI denote the accuracies on matched and mis-matched sections respectively. }
\end{table}

\begin{table*}[tbp]
\centering
\resizebox{0.8\textwidth}{!}{

\begin{tabular}{lccccc}
\toprule
\textbf{Model} & \textbf{\# Layers} & \textbf{Hidden size} & \textbf{Feed-forward size} & \textbf{\# Parameters} & \textbf{Relative size} \\ \midrule
BERT$_{\text{\tt BASE}}$ (teacher) & 12 & 768 & 3072 & 108M & 100\%\\ \midrule
T6           & 6 & 768 & 3072 & 65M & 60\% \\
T3           & 3 & 768 & 3072 & 44M & 41\% \\
T3-small & 3 & 384 & 1536 & 17M & 16\% \\
T4-tiny & 4 & 312 & 1200 & 14M & 13\% \\
BiGRU & 1 & 768 & -& 31M & 29\% \\ \bottomrule
\end{tabular}
}
\caption{\label{model-configurations}  Model sizes of teacher and students. The number of parameters includes embeddings but does not include output layers.}
\end{table*}

\textbf{Training settings}. To keep experiments simple, we directly distill the teacher model that has been trained on the task, while we do not perform task-irrelevant language modeling distillation in advance. 
The number of epochs ranges from 30 to 60, and the learning rate of students is 1e-4 for all distillation experiments. 

\textbf{Distillation settings}.
Temperature is set to 8 for all experiments. We add intermediate losses uniformly distributed among all the layers between teacher and student (except BiGRU). The loss functions we choose are \texttt{hidden\_mse} loss which computes the mean square loss between two hidden states, and \texttt{NST} loss which is an effective method in CV. 
In Figure \ref{distill-config} we show an example of distillation configuration for distilling BERT$_{\text{\tt BASE}}$ to a T4-tiny.  
Since their hidden sizes are different, we use \texttt{proj} option to add linear layers to match the dimensions. The linear layers will be trained together with the student automatically.
We experiment with two kinds of distillers:  \texttt{GeneralDistiller} and \texttt{MultiTeacherDistiller} .

\subsection{Results on English Datasets}
We list the public results (DistilBERT and TinyBERT) and our distillation results obtained by \texttt{GeneralDistiller} in Table \ref{distiilation-results-english}.  We have the following observations.

First, teachers can be distilled to T6 models with minor losses in performance.
All the T6 models achieve 99\% performance of the teachers, higher than the DistilBERT.

Second, T4-tiny outperforms TinyBERT though they share the same structure. This is attributed to the NST losses in the distillation configuration.  This result proves the effectiveness of applying KD method developed in  CV on NLP tasks.

Third, although T4-tiny has less parameters than T3-small,  T4-tiny outperforms T3-small in most cases. It may be a hint that narrow-and-deep models are better than wide-and-shallow models. 

Finally, data augmentation (DA) is critical. For the experiments in the last line in Table \ref{distiilation-results-english}, we use additional datasets during distillation: a subset of NewsQA \cite{trischler-etal-2017-newsqa} training set is used in SQuAD; passages from the HotpotQA \cite{yang-etal-2018-hotpotqa} training set is used in CoNLL-2003. The augmentation datasets significantly improve the performance, especially when the size of the training set is small, like CoNLL-2003.

We next show the effectiveness of \texttt{MultiTeacherDistiller}, which distills an ensemble of teachers to a single student model. 
For each task, we train three  BERT$_{\text{\tt BASE}}$  teacher models with different seeds. The student is also a BERT$_{\text{\tt BASE}}$ model. The temperature is set to 8, and intermediate losses are not used. 
As Table \ref{multi-teacher-distillation} shows, for each task, the student achieves the best performance,  even higher than the ensemble result.

\begin{table}[t!]
\resizebox{\columnwidth}{!}{
\begin{tabular}{lccccc}
\toprule
\multirow{2}{*}{\textbf{Model}} & \multicolumn{2}{c}{\textbf{MNLI}} & \multicolumn{2}{c}{\textbf{SQuAD}} &{\textbf{CoNLL-2003}} \\ 
 & m & mm & EM & F1  & F1       \\ \midrule
Teacher 1 & 83.6 & 84.0 & 81.1 & 88.6 &91.2 \\ 
Teacher 2 & 83.6 & 84.2 & 81.2 & 88.5 & 90.8 \\
Teacher 3  & 83.7 & 83.8 & 81.2 & 88.7 & 91.3 \\ \midrule
Ensemble  & 84.3 & 84.7 & 82.3 & 89.4 & 91.5 \\ \midrule
Student    & \textbf{84.8} & \textbf{85.3} & \textbf{83.5} & \textbf{90.0} & \textbf{91.6}\\ \bottomrule
\end{tabular}
}
\caption{\label{multi-teacher-distillation} Results of multi-teacher distillation. 
All the models are BERT$_{\text{\tt BASE}}$.
Different teachers are trained with different random seeds. 
For each task, the ensemble is the average of three teachers' results.}
\end{table}

\section{Results on Chinese Datasets}

The results on Chinese datasets are presented in Table \ref{distillation-results-chinese}. 
We notice that T4-tiny still outperforms T3-small on all tasks, which is consistent with their performance on English tasks.
In the experiments with DA,  CMRC 2018 and DRCD take each other's dataset as data augmentation.
We observe that since CMRC 2018 has a relatively small training set, 
DA has a much more significant effect.

\begin{table}[tbp]
    \centering
\resizebox{\columnwidth}{!}{

    \begin{tabular}{lcccccc}
    \toprule
    \multirow{2}{*}{\textbf{Model}} & {\textbf{XNLI}} & {\textbf{LCQMC}} & \multicolumn{2}{c}{\textbf{CMRC 2018}} & \multicolumn{2}{c}{\textbf{DRCD}} \\ 
     & Acc & Acc & EM & F1 & EM & F1        \\ \midrule
   RoBERTa-wwm  & 79.9 & 89.4 & 68.8 & 86.4 & 86.5 & 92.5\\ \midrule
   T3 & 78.4 & 89.0 & 63.4 & 82.4 & 76.7 & 85.2 \\
   \ \ +DA & - & - & 66.4 & 84.2 & 78.2 & 86.4 \\
   T3-small & 76.0 & 88.1 & 46.1 & 71.0     & 71.4 & 82.2 \\
   \ \ +DA & - & - & 58.0 & 79.3 & 75.8 &  84.8 \\
    T4-tiny  & 76.2 & 88.4 & 54.3 & 76.8   & 75.5 & 84.9\\
    \ \ +DA & -& - & 61.8 & 81.8 & 77.3 & 86.1 \\ \bottomrule
    
    \end{tabular}
   }
    \caption{\label{distillation-results-chinese} Development set results for the teacher and various students on Chinese tasks.}
\end{table}

\section{Conclusion and Future Work}
In this paper, we present TextBrewer, a flexible PyTorch-based distillation toolkit for NLP research and applications. 
TextBrewer provides rich customization options for users to compare different distillation methods and build their strategies. 
We have conducted a series of experiments. The results show that the distilled models can achieve state-of-the-art results with simple settings.

TextBrewer also has its limitations. For example, its usability in generation tasks such as machine translation has not been tested.
We will keep adding more examples and tests to expand TextBrewer's scope of application. 

Apart from the distillation strategies, the model structure also affects the performance. 
In the future, we aim to integrate neural architecture search into the toolkit to automate the searching for model structures.

\section*{Acknowledgments}
We would like to thank all anonymous reviewers for their valuable comments on our work. 
This work was supported by the National Natural Science Foundation of China (NSFC) via grant 61976072, 61632011, and 61772153.

\bibliography{my_nlp}

\begin{thebibliography}{30}
\expandafter\ifx\csname natexlab\endcsname\relax\def\natexlab#1{#1}\fi

\bibitem[{Clark et~al.(2019)Clark, Luong, Khandelwal, Manning, and
  Le}]{clark-etal-2019-bam}
Kevin Clark, Minh-Thang Luong, Urvashi Khandelwal, Christopher~D. Manning, and
  Quoc~V. Le. 2019.
\newblock \href {https://doi.org/10.18653/v1/P19-1595} {{BAM}! born-again
  multi-task networks for natural language understanding}.
\newblock In \emph{Proceedings of the 57th Annual Meeting of the Association
  for Computational Linguistics}, pages 5931--5937, Florence, Italy.
  Association for Computational Linguistics.

\bibitem[{Conneau et~al.(2018)Conneau, Rinott, Lample, Williams, Bowman,
  Schwenk, and Stoyanov}]{conneau-etal-2018-xnli}
Alexis Conneau, Ruty Rinott, Guillaume Lample, Adina Williams, Samuel Bowman,
  Holger Schwenk, and Veselin Stoyanov. 2018.
\newblock \href {https://doi.org/10.18653/v1/D18-1269} {{XNLI}: Evaluating
  cross-lingual sentence representations}.
\newblock In \emph{Proceedings of the 2018 Conference on Empirical Methods in
  Natural Language Processing}, pages 2475--2485, Brussels, Belgium.
  Association for Computational Linguistics.

\bibitem[{Cui et~al.(2019{\natexlab{a}})Cui, Che, Liu, Qin, Yang, Wang, and
  Hu}]{DBLP:journals/corr/abs-1906-08101}
Yiming Cui, Wanxiang Che, Ting Liu, Bing Qin, Ziqing Yang, Shijin Wang, and
  Guoping Hu. 2019{\natexlab{a}}.
\newblock \href {http://arxiv.org/abs/1906.08101} {Pre-training with whole word
  masking for chinese {BERT}}.
\newblock \emph{CoRR}, abs/1906.08101.

\bibitem[{Cui et~al.(2019{\natexlab{b}})Cui, Liu, Che, Xiao, Chen, Ma, Wang,
  and Hu}]{cui-emnlp2019-cmrc2018}
Yiming Cui, Ting Liu, Wanxiang Che, Li~Xiao, Zhipeng Chen, Wentao Ma, Shijin
  Wang, and Guoping Hu. 2019{\natexlab{b}}.
\newblock \href {https://doi.org/10.18653/v1/D19-1600} {A span-extraction
  dataset for {C}hinese machine reading comprehension}.
\newblock In \emph{Proceedings of the 2019 Conference on Empirical Methods in
  Natural Language Processing and the 9th International Joint Conference on
  Natural Language Processing (EMNLP-IJCNLP)}, pages 5886--5891, Hong Kong,
  China. Association for Computational Linguistics.

\bibitem[{Devlin et~al.(2019)Devlin, Chang, Lee, and
  Toutanova}]{devlin-etal-2019-bert}
Jacob Devlin, Ming-Wei Chang, Kenton Lee, and Kristina Toutanova. 2019.
\newblock \href {https://doi.org/10.18653/v1/N19-1423} {{BERT}: Pre-training of
  deep bidirectional transformers for language understanding}.
\newblock In \emph{Proceedings of the 2019 Conference of the North {A}merican
  Chapter of the Association for Computational Linguistics: Human Language
  Technologies, Volume 1 (Long and Short Papers)}, pages 4171--4186,
  Minneapolis, Minnesota. Association for Computational Linguistics.

\bibitem[{Huang and Wang(2017)}]{DBLP:journals/corr/HuangW17a}
Zehao Huang and Naiyan Wang. 2017.
\newblock \href {http://arxiv.org/abs/1707.01219} {Like what you like:
  Knowledge distill via neuron selectivity transfer}.
\newblock \emph{CoRR}, abs/1707.01219.

\bibitem[{Jiao et~al.(2019)Jiao, Yin, Shang, Jiang, Chen, Li, Wang, and
  Liu}]{DBLP:journals/corr/abs-1909-10351}
Xiaoqi Jiao, Yichun Yin, Lifeng Shang, Xin Jiang, Xiao Chen, Linlin Li, Fang
  Wang, and Qun Liu. 2019.
\newblock \href {http://arxiv.org/abs/1909.10351} {Tinybert: Distilling {BERT}
  for natural language understanding}.
\newblock \emph{CoRR}, abs/1909.10351.

\bibitem[{Kim and Rush(2016)}]{kim-rush-2016-sequence}
Yoon Kim and Alexander~M. Rush. 2016.
\newblock \href {https://doi.org/10.18653/v1/D16-1139} {Sequence-level
  knowledge distillation}.
\newblock In \emph{Proceedings of the 2016 Conference on Empirical Methods in
  Natural Language Processing}, pages 1317--1327, Austin, Texas. Association
  for Computational Linguistics.

\bibitem[{Lan et~al.(2019)Lan, Chen, Goodman, Gimpel, Sharma, and
  Soricut}]{albert}
Zhenzhong Lan, Mingda Chen, Sebastian Goodman, Kevin Gimpel, Piyush Sharma, and
  Radu Soricut. 2019.
\newblock \href {http://arxiv.org/abs/1909.11942} {{ALBERT:} {A} lite {BERT}
  for self-supervised learning of language representations}.
\newblock \emph{CoRR}, abs/1909.11942.

\bibitem[{Liu et~al.(2019{\natexlab{a}})Liu, He, Chen, and
  Gao}]{DBLP:journals/corr/abs-1904-09482}
Xiaodong Liu, Pengcheng He, Weizhu Chen, and Jianfeng Gao. 2019{\natexlab{a}}.
\newblock \href {http://arxiv.org/abs/1904.09482} {Improving multi-task deep
  neural networks via knowledge distillation for natural language
  understanding}.
\newblock \emph{CoRR}, abs/1904.09482.

\bibitem[{Liu et~al.(2018)Liu, Chen, Deng, Zeng, Chen, Li, and
  Tang}]{liu-etal-2018-lcqmc}
Xin Liu, Qingcai Chen, Chong Deng, Huajun Zeng, Jing Chen, Dongfang Li, and
  Buzhou Tang. 2018.
\newblock \href {https://www.aclweb.org/anthology/C18-1166} {{LCQMC}:a
  large-scale {C}hinese question matching corpus}.
\newblock In \emph{Proceedings of the 27th International Conference on
  Computational Linguistics}, pages 1952--1962, Santa Fe, New Mexico, USA.
  Association for Computational Linguistics.

\bibitem[{Liu et~al.(2019{\natexlab{b}})Liu, Ott, Goyal, Du, Joshi, Chen, Levy,
  Lewis, Zettlemoyer, and Stoyanov}]{DBLP:journals/corr/abs-1907-11692}
Yinhan Liu, Myle Ott, Naman Goyal, Jingfei Du, Mandar Joshi, Danqi Chen, Omer
  Levy, Mike Lewis, Luke Zettlemoyer, and Veselin Stoyanov. 2019{\natexlab{b}}.
\newblock \href {http://arxiv.org/abs/1907.11692} {Roberta: {A} robustly
  optimized {BERT} pretraining approach}.
\newblock \emph{CoRR}, abs/1907.11692.

\bibitem[{Radford(2018)}]{gpt}
Alec Radford. 2018.
\newblock Improving language understanding by generative pre-training.

\bibitem[{Rajpurkar et~al.(2016)Rajpurkar, Zhang, Lopyrev, and
  Liang}]{rajpurkar-etal-2016-squad}
Pranav Rajpurkar, Jian Zhang, Konstantin Lopyrev, and Percy Liang. 2016.
\newblock \href {https://doi.org/10.18653/v1/D16-1264} {{SQ}u{AD}: 100,000+
  questions for machine comprehension of text}.
\newblock In \emph{Proceedings of the 2016 Conference on Empirical Methods in
  Natural Language Processing}, pages 2383--2392, Austin, Texas. Association
  for Computational Linguistics.

\bibitem[{Sanh et~al.(2019)Sanh, Debut, Chaumond, and
  Wolf}]{sanh2019distilbert}
Victor Sanh, Lysandre Debut, Julien Chaumond, and Thomas Wolf. 2019.
\newblock \href {http://arxiv.org/abs/1910.01108} {Distilbert, a distilled
  version of {BERT:} smaller, faster, cheaper and lighter}.
\newblock \emph{CoRR}, abs/1910.01108.

\bibitem[{Shao et~al.(2018)Shao, Liu, Lai, Tseng, and
  Tsai}]{DBLP:journals/corr/abs-1806-00920}
Chih{-}Chieh Shao, Trois Liu, Yuting Lai, Yiying Tseng, and Sam Tsai. 2018.
\newblock \href {http://arxiv.org/abs/1806.00920} {{DRCD:} a chinese machine
  reading comprehension dataset}.
\newblock \emph{CoRR}, abs/1806.00920.

\bibitem[{Sun et~al.(2019{\natexlab{a}})Sun, Cheng, Gan, and
  Liu}]{sun-etal-2019-patient}
Siqi Sun, Yu~Cheng, Zhe Gan, and Jingjing Liu. 2019{\natexlab{a}}.
\newblock \href {https://doi.org/10.18653/v1/D19-1441} {Patient knowledge
  distillation for {BERT} model compression}.
\newblock In \emph{Proceedings of the 2019 Conference on Empirical Methods in
  Natural Language Processing and the 9th International Joint Conference on
  Natural Language Processing (EMNLP-IJCNLP)}, pages 4323--4332, Hong Kong,
  China. Association for Computational Linguistics.

\bibitem[{Sun et~al.(2019{\natexlab{b}})Sun, Wang, Li, Feng, Chen, Zhang, Tian,
  Zhu, Tian, and Wu}]{baidu-ernie-1}
Yu~Sun, Shuohuan Wang, Yu{-}Kun Li, Shikun Feng, Xuyi Chen, Han Zhang, Xin
  Tian, Danxiang Zhu, Hao Tian, and Hua Wu. 2019{\natexlab{b}}.
\newblock \href {http://arxiv.org/abs/1904.09223} {{ERNIE:} enhanced
  representation through knowledge integration}.
\newblock \emph{CoRR}, abs/1904.09223.

\bibitem[{Sun et~al.(2019{\natexlab{c}})Sun, Wang, Li, Feng, Tian, Wu, and
  Wang}]{baidu-ernie-2}
Yu~Sun, Shuohuan Wang, Yu{-}Kun Li, Shikun Feng, Hao Tian, Hua Wu, and Haifeng
  Wang. 2019{\natexlab{c}}.
\newblock \href {http://arxiv.org/abs/1907.12412} {{ERNIE} 2.0: {A} continual
  pre-training framework for language understanding}.
\newblock \emph{CoRR}, abs/1907.12412.

\bibitem[{Tan et~al.(2019)Tan, Ren, He, Qin, Zhao, and
  Liu}]{DBLP:conf/iclr/TanRHQZL19}
Xu~Tan, Yi~Ren, Di~He, Tao Qin, Zhou Zhao, and Tie{-}Yan Liu. 2019.
\newblock \href {https://openreview.net/forum?id=S1gUsoR9YX} {Multilingual
  neural machine translation with knowledge distillation}.
\newblock In \emph{7th International Conference on Learning Representations,
  {ICLR} 2019, New Orleans, LA, USA, May 6-9, 2019}.

\bibitem[{Tang et~al.(2019)Tang, Lu, and Lin}]{tang-etal-2019-natural}
Raphael Tang, Yao Lu, and Jimmy Lin. 2019.
\newblock \href {https://doi.org/10.18653/v1/D19-6122} {Natural language
  generation for effective knowledge distillation}.
\newblock In \emph{Proceedings of the 2nd Workshop on Deep Learning Approaches
  for Low-Resource NLP (DeepLo 2019)}, pages 202--208, Hong Kong, China.
  Association for Computational Linguistics.

\bibitem[{Tjong Kim~Sang and
  De~Meulder(2003)}]{tjong-kim-sang-de-meulder-2003-introduction}
Erik~F. Tjong Kim~Sang and Fien De~Meulder. 2003.
\newblock \href {https://www.aclweb.org/anthology/W03-0419} {Introduction to
  the {C}o{NLL}-2003 shared task: Language-independent named entity
  recognition}.
\newblock In \emph{Proceedings of the Seventh Conference on Natural Language
  Learning at {HLT}-{NAACL} 2003}, pages 142--147.

\bibitem[{Trischler et~al.(2017)Trischler, Wang, Yuan, Harris, Sordoni,
  Bachman, and Suleman}]{trischler-etal-2017-newsqa}
Adam Trischler, Tong Wang, Xingdi Yuan, Justin Harris, Alessandro Sordoni,
  Philip Bachman, and Kaheer Suleman. 2017.
\newblock \href {https://doi.org/10.18653/v1/W17-2623} {{N}ews{QA}: A machine
  comprehension dataset}.
\newblock In \emph{Proceedings of the 2nd Workshop on Representation Learning
  for {NLP}}, pages 191--200, Vancouver, Canada. Association for Computational
  Linguistics.

\bibitem[{Wang et~al.(2019)Wang, Singh, Michael, Hill, Levy, and
  Bowman}]{DBLP:conf/iclr/WangSMHLB19}
Alex Wang, Amanpreet Singh, Julian Michael, Felix Hill, Omer Levy, and
  Samuel~R. Bowman. 2019.
\newblock \href {https://openreview.net/forum?id=rJ4km2R5t7} {{GLUE:} {A}
  multi-task benchmark and analysis platform for natural language
  understanding}.
\newblock In \emph{7th International Conference on Learning Representations,
  {ICLR} 2019, New Orleans, LA, USA, May 6-9, 2019}.

\bibitem[{Wen et~al.(2019)Wen, Lai, and
  Qian}]{DBLP:journals/corr/abs-1911-07471}
Tiancheng Wen, Shenqi Lai, and Xueming Qian. 2019.
\newblock \href {http://arxiv.org/abs/1911.07471} {Preparing lessons: Improve
  knowledge distillation with better supervision}.
\newblock \emph{CoRR}, abs/1911.07471.

\bibitem[{Yang et~al.(2019)Yang, Dai, Yang, Carbonell, Salakhutdinov, and
  Le}]{DBLP:journals/corr/abs-1906-08237}
Zhilin Yang, Zihang Dai, Yiming Yang, Jaime~G. Carbonell, Ruslan Salakhutdinov,
  and Quoc~V. Le. 2019.
\newblock \href {http://arxiv.org/abs/1906.08237} {Xlnet: Generalized
  autoregressive pretraining for language understanding}.
\newblock \emph{CoRR}, abs/1906.08237.

\bibitem[{Yang et~al.(2018)Yang, Qi, Zhang, Bengio, Cohen, Salakhutdinov, and
  Manning}]{yang-etal-2018-hotpotqa}
Zhilin Yang, Peng Qi, Saizheng Zhang, Yoshua Bengio, William Cohen, Ruslan
  Salakhutdinov, and Christopher~D. Manning. 2018.
\newblock \href {https://doi.org/10.18653/v1/D18-1259} {{H}otpot{QA}: A dataset
  for diverse, explainable multi-hop question answering}.
\newblock In \emph{Proceedings of the 2018 Conference on Empirical Methods in
  Natural Language Processing}, pages 2369--2380, Brussels, Belgium.
  Association for Computational Linguistics.

\bibitem[{Yim et~al.(2017)Yim, Joo, Bae, and Kim}]{DBLP:conf/cvpr/YimJBK17}
Junho Yim, Donggyu Joo, Ji{-}Hoon Bae, and Junmo Kim. 2017.
\newblock \href {https://doi.org/10.1109/CVPR.2017.754} {A gift from knowledge
  distillation: Fast optimization, network minimization and transfer learning}.
\newblock In \emph{2017 {IEEE} Conference on Computer Vision and Pattern
  Recognition, {CVPR} 2017, Honolulu, HI, USA, July 21-26, 2017}, pages
  7130--7138.

\bibitem[{Zhao et~al.(2019)Zhao, Gupta, Song, and
  Zhou}]{DBLP:journals/corr/abs-1909-11687}
Sanqiang Zhao, Raghav Gupta, Yang Song, and Denny Zhou. 2019.
\newblock \href {http://arxiv.org/abs/1909.11687} {Extreme language model
  compression with optimal subwords and shared projections}.
\newblock \emph{CoRR}, abs/1909.11687.

\bibitem[{Zmora et~al.(2018)Zmora, Jacob, Zlotnik, Elharar, and
  Novik}]{neta_zmora_2018_1297430}
Neta Zmora, Guy Jacob, Lev Zlotnik, Bar Elharar, and Gal Novik. 2018.
\newblock \href {https://doi.org/10.5281/zenodo.1297430} {Neural network
  distiller}.

\end{thebibliography}
\bibliographystyle{acl_natbib}

\end{document}